\documentclass{article} % For LaTeX2e
\usepackage{nips14submit_e,times}
\usepackage{hyperref}
\usepackage{url}

\usepackage{algorithm,algorithmic,color}
\usepackage{float}
\usepackage{amssymb}
\usepackage{amsmath}
\usepackage{amsthm}
\usepackage{subfig}
\usepackage{multirow}
\usepackage{graphicx}
\usepackage{color}
\usepackage{stmaryrd}
\usepackage{wrapfig}

\definecolor{darkblue}{rgb}{0,0.05,0.35}
\definecolor{darkgreen}{rgb}{0,0.6,0}

\newcommand{\fig}[1]{Fig.~\ref{fig:#1}}
\newcommand{\tab}[1]{Table~\ref{tab:#1}}
\newcommand{\secc}[1]{Section~\ref{sec:#1}}

\usepackage{bm,array}

\newlength{\cifarwidth}
\usepackage[compact]{titlesec}
\titlespacing{\section}{0pt}{2ex}{1ex}
\titlespacing{\subsection}{0pt}{1ex}{0ex}
\titlespacing{\subsubsection}{0pt}{0.5ex}{0ex}

\newcolumntype{C}{>{\centering\arraybackslash}p{6em}}

\usepackage{bm}
\newif\ifMIT
\MITfalse
\usepackage{amsmath}
\usepackage{amsfonts}
\usepackage{amsthm}
\def\1{\bm{1}}

\def\vx{{\bm{x}}}

\def\vz{{\bm{z}}}
\ifMIT
\def\gamma{{\boldmath{\vgamma}}}
\def\vtheta{{\boldmath{\theta}}}

\else

\def\vgamma{{\bm{\gamma}}}
\def\vtheta{{\bm{\theta}}}

\fi

\ifMIT

\else

\fi

\DeclareMathAlphabet{\mathsfit}{\encodingdefault}{\sfdefault}{m}{sl}
\SetMathAlphabet{\mathsfit}{bold}{\encodingdefault}{\sfdefault}{bx}{n}

\newcommand{\pdata}{p_{\rm{data}}}
\newcommand{\pmodel}{p_{\rm{model}}}

\newcommand{\E}{\mathbb{E}}

\ifMIT
\else

\fi

\title{Improved Techniques for Training GANs}

\author{Tim Salimans \\
\texttt{tim@openai.com} \\
\And
Ian Goodfellow\\
\texttt{ian@openai.com} \\
\And
Wojciech Zaremba\\
\texttt{woj@openai.com} \\
\And
Vicki  Cheung\\
\texttt{vicki@openai.com}
\And
Alec  Radford\\
\texttt{alec.radford@gmail.com}
\And
Xi Chen\\
\texttt{peter@openai.com}
}

\newcommand{\f}{\mathbf{f}}

\newcommand{\X}{\mathbf{X}}

\nipsfinalcopy % Uncomment for camera-ready version

\begin{document}
\setlength{\lineskiplimit}{0pt}
\setlength{\lineskip}{0pt}
\setlength{\abovedisplayskip}{2pt}
\setlength{\belowdisplayskip}{0pt}

\maketitle

\begin{abstract}
We present a variety of new architectural features and training procedures that we apply
to the generative adversarial networks (GANs) framework.
We focus on two applications of GANs: semi-supervised learning, and the generation of
images that humans find visually realistic.
Unlike most work on generative models, our primary goal is not to train a model that assigns
high likelihood to test data, nor do we require the model to be able to learn well without using any
labels. Using our new techniques, we achieve state-of-the-art results in semi-supervised classification
on MNIST, CIFAR-10 and SVHN.
The generated images are of high quality as confirmed by a visual Turing test: 
our model generates MNIST samples that humans cannot distinguish from real data,
and CIFAR-10 samples that yield a human error rate of $21.3\%$.
We also present ImageNet samples with unprecedented resolution and show that
our methods enable the model to learn recognizable features of ImageNet classes.
\end{abstract}

%\tableofcontents

\section{Introduction}
Generative adversarial networks~\cite{goodfellow2014generative} (GANs) are a class of methods for learning generative models
based on game theory. The goal of GANs is to train a generator network $G(\vz; \vtheta^{(G)})$ that produces samples from the
data distribution, $\pdata(\vx)$, by transforming vectors of noise $\vz$ as $\vx = G(\vz; \vtheta^{(G)})$. The training signal for $G$ is provided by a discriminator network $D(\vx)$ that is trained to distinguish samples from the generator distribution $\pmodel(\vx)$ from real data. The generator network $G$ in turn is then trained to fool the discriminator into accepting its outputs as being real.

Recent applications of GANs have shown that they can produce excellent samples \cite{denton2015deep, radford2015unsupervised}.
However, training GANs requires finding a Nash equilibrium of a non-convex game
with continuous, high-dimensional parameters.
GANs are typically trained using gradient descent techniques that are designed to
find a low value of a cost function, rather than to find the Nash equilibrium of
a game.
When used to seek for a Nash equilibrium, these algorithms 
may fail to converge \cite{goodfellow2014distinguishability}.

In this work, we introduce several techniques intended to encourage convergence of
the GANs game.
These techniques are motivated by a heuristic understanding of the non-convergence
problem.
They lead to improved semi-supervised learning peformance and improved sample generation.
We hope that some of them may form the basis for future work, providing formal guarantees
of convergence.

All code and hyperparameters may be found at:
\url{https://github.com/openai/improved_gan}

\section{Related work}
\label{subsec:previous}
% This info is already in the intro:
% Generative adversarial networks is a recently developed class of generative models~\cite{goodfellow2014generative} that has shown to be able to achieve amazing visual sample quality, but that can also suffer from instability problems.
Several recent papers focus on improving the stability of training and the
resulting perceptual quality of GAN
samples~\cite{denton2015deep, radford2015unsupervised, im2016generating, yoo2016pixel}.
We build on some of these techniques in this work.
For instance, we use some of the ``DCGAN'' architectural innovations proposed
in Radford et al.~\cite{radford2015unsupervised}, as discussed below.
% We can postpone the detail
% such as using batch-normalization~\cite{ioffe2015batch} in the generator network, and strided convolutions in the discriminator.

One of our proposed techniques, {\em feature matching}, discussed in Sec. \ref{sec:feature},
is similar in spirit to approaches that use maximum mean discrepancy \cite{gretton2005measuring, fukumizu2007kernel, smola2007hilbert}
to train generator networks \cite{Li-et-al-2015,dziugaite2015training}.
Another of our proposed techniques, {\em minibatch features}, is based in part on ideas used
for batch normalization~\cite{ioffe2015batch}, while our proposed 
 {\em virtual batch normalization} is a direct extension of batch normalization.

% Cut to save space.
%
% The main purpose of this work, however, is in developing methods for strong semi-supervised learning~\cite{zhu2005semi}.
% Many classical semi-supervised methods rely on propagating labels over unlabeled
% data~\cite{zhu2002learning, blum2001learning}.
% This approach is quite successful in some settings, however it does not work
% well when the number of labeled examples is very small.
%
% Don't we make this same assumption?
% 
% Moreover, it relies on the restrictive assumption that unlabeled examples
% belong to the same domain as the training data.

One of the primary goals of this work is to improve the effectiveness of generative
adversarial networks for semi-supervised learning (improving the performance of
a supervised task, in this case, classification, by learning on additional
unlabeled examples).
Like many deep generative models, GANs have previously been applied to
semi-supervised learning \cite{sutskever2015towards,springenberg2015unsupervised}, and our
work can be seen as a continuation and refinement of this effort.

\section{Toward Convergent GAN Training}
Training GANs consists in finding a Nash equilibrium to a two-player non-cooperative game.
Each player wishes to minimize its own cost function, $J^{(D)}(\vtheta^{(D)}, \vtheta^{(G)})$
for the discriminator
and $J^{(G)}(\vtheta^{(D)}, \vtheta^{(G)})$ for the generator.
A Nash equilibirum is a point $(\vtheta^{(D)}, \vtheta^{(G)})$ such that $J^{(D)}$ is
at a minimum with respect to $\vtheta^{(D)}$ and $J^{(G)}$ is at a minimum with respect to $\vtheta^{(G)}$.
Unfortunately, finding Nash equilibria is a very difficult problem. Algorithms exist for specialized cases, but we are not aware of any that are feasible
to apply to the GAN game, where the cost functions are non-convex, the parameters are
continuous, and the parameter space is extremely high-dimensional.

The idea that a Nash equilibrium occurs when each player has minimal cost seems
to intuitively motivate the idea of using traditional gradient-based minimization
techniques to minimize each player's cost simultaneously.
Unfortunately, a modification to $\vtheta^{(D)}$ that reduces $J^{(D)}$ can increase
$J^{(G)}$, and a modification to $\vtheta^{(G)}$ that reduces $J^{(G)}$ can increase
$J^{(D)}$.
Gradient descent thus fails to converge for many games.
For example, when one player minimizes $xy$ with respect to $x$ and another player
minimizes $-xy$ with respect to $y$, gradient descent enters a stable orbit,
rather than converging to $x=y=0$, the desired equilibrium point \cite{Goodfellow-et-al-2016-Book}.
Previous approaches to GAN training have thus applied gradient descent on each player's
cost simultaneously, despite the lack of guarantee that this procedure will converge.
We introduce the following techniques that are heuristically motivated to encourage
convergence:

\subsection{Feature matching}
\label{sec:feature}
Feature matching addresses the instability of GANs
by specifying a new objective for the generator that prevents it from overtraining on the current discriminator. Instead of directly maximizing the output of the discriminator, the new objective requires the generator to generate data that matches the statistics of the real data, where we use the discriminator only to specify the statistics that we think are worth matching. Specifically, we train the generator to match the expected value of the features on an intermediate layer of the discriminator. This is a natural choice of statistics for the generator to match, since by training the discriminator we ask it to find those features that are most discriminative of real data versus data generated by the current model. %, i.e.\ those features which currently have the highest signal-to-noise ratio for distinguishing real data from generated data. %After the generator has learned to match these features, the discriminator will evolve to find another set of features which are most discriminative against the new generator. By optimizing $G$ and $D$ in tandem, we thus effectively have the generator continuously minimize the worst statistical mismatch between the generated data and real data. An important advantage of this strategy is that this objective can still be optimized even when the discriminator is completely untrained or when it completely distinguishes real data from generated. In contrast, conventional GAN training requires the training of the generator and discriminator to happen in lock-step in order to avoid instability. 

Letting $\f(\vx)$ denote activations on an intermediate layer of the discriminator,
our new objective for the generator is defined as:
$||\E_{\vx \sim p_{\text{data}}}\f(\vx) - \E_{\vz \sim p_{\vz}(\vz)}\f(G(\vz))||_{2}^{2}$.
The discriminator, and hence $\f(\vx)$, are trained in the usual way.
As with regular GAN training, the objective has a fixed point where $G$
exactly matches the distribution of training data.
We have no guarantee of reaching this fixed point in practice, but our empirical
results indicate that feature matching is indeed effective in situations where
regular GAN becomes unstable.

\subsection{Minibatch discrimination}
\label{sec:minibatch}
One of the main failure modes for GAN is for the generator to collapse to a parameter setting where it
always emits the same point.
When collapse to a single mode is imminent, the gradient of the
discriminator may point in similar directions for many similar points.
Because the discriminator processes each example independently, there is
no coordination between its gradients, and thus no mechanism to tell the
outputs of the generator to become more dissimilar to each other.
Instead, all outputs race toward a single point that the discriminator
currently believes is highly realistic.
After collapse has occurred, the discriminator learns that this single
point comes from the generator, but gradient descent is unable to separate
the identical outputs. The gradients of the discriminator then push the
single point produced by the generator around space forever, and the
algorithm cannot converge to a distribution with the correct amount of entropy.
% Ian doesn't agree with the next part. Usually the discriminator learns
% to detect these problems fairly quickly, which causes the mode to move
% around.
% 
% Such a collapse is difficult for the discriminator to detect as it processes
% each example separately: The generated samples might individually look very
% much like real data (they might be perfect copies), even though they are in
% fact all identical and do not cover the data distribution well at all.
% For the human experimenter such a failure is immediately obvious because we look at multiple generator samples combined.
An obvious strategy to avoid this type of failure is to allow the discriminator to look at multiple data examples in combination, and perform what we call \emph{minibatch discrimination}.

\begin{wrapfigure}[13]{r}{0.4\textwidth}
	\centering
	\includegraphics[width=0.38\textwidth]{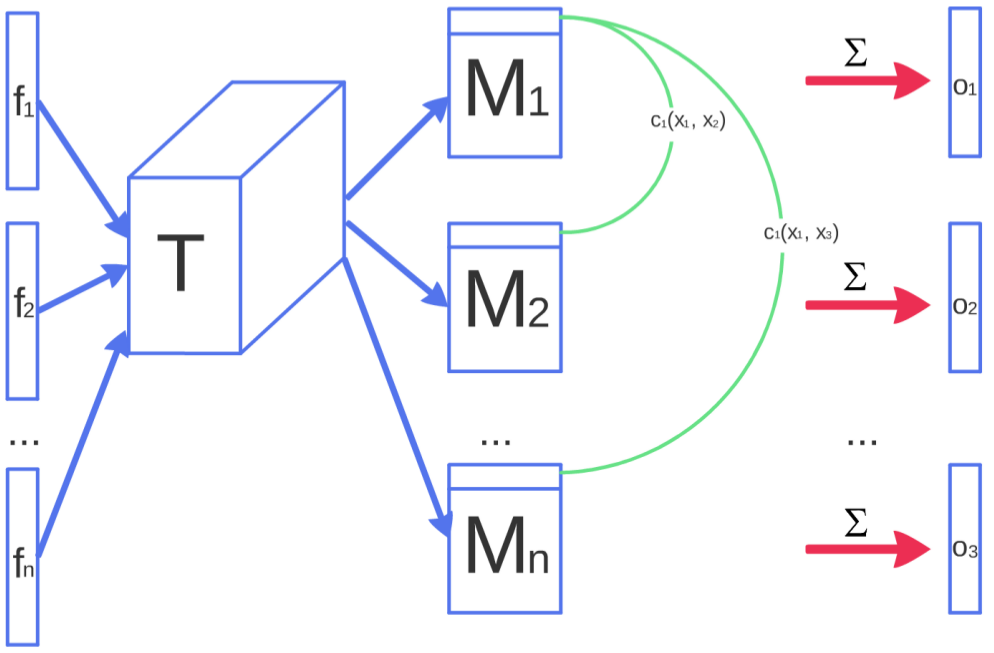}
	\caption{Figure sketches how minibatch discrimination works. Features $\f(\vx_{i})$ from sample $\vx_{i}$ are multiplied
    through a tensor $T$, and cross-sample distance is computed.}
\label{fig:dist}
\end{wrapfigure}

The concept of minibatch discrimination is quite general: any discriminator model that looks at multiple examples in combination, rather than in isolation, could potentially help avoid collapse of the generator. In fact, the successful application of batch normalization in the discriminator by Radford et al.~\cite{radford2015unsupervised} is well explained from this perspective. So far, however,  we have restricted our experiments to models that explicitly aim to identify generator samples that are particularly close together. One successful specification for modelling the \emph{closeness} between examples in a minibatch is as follows: Let $\f(\vx_{i}) \in \mathbb{R}^A$ denote a vector of features for input $\vx_{i}$, produced by some intermediate layer in the discriminator. We then multiply the vector $\f(\vx_{i})$ by a tensor $T \in \mathbb{R}^{A \times B \times C}$, which results in a matrix $M_i \in \mathbb{R}^{B \times C}$. We then compute the $L_1$-distance between the rows of the resulting matrix $M_i$ across samples $i \in \{1, 2, \dots, n\}$ and apply a negative exponential (\fig{dist}): $c_b(\vx_{i}, \vx_{j}) = \exp(-||M_{i, b} - M_{j, b}||_{L_1}) \in \mathbb{R}$.
The output $o(x_{i})$ for this \emph{minibatch layer} for a sample $\vx_{i}$ is then defined as the sum of the $c_b(\vx_{i},\vx_{j})$'s to all other samples:
\begin{align*}
    o(\vx_{i})_b &= \sum_{j=1}^{n} c_b(\vx_{i},\vx_{j}) \in \mathbb{R} \\
    o(\vx_{i}) &= \Big[ o(\vx_{i})_1, o(\vx_{i})_2, \dots, o(\vx_{i})_B \Big] \in \mathbb{R}^B \\
    o(\X) &\in \mathbb{R}^{n \times B}
\end{align*}

Next, we concatenate the output $o(\vx_{i})$ of the minibatch layer with the intermediate features $\f(\vx_{i})$ that were its input, and we feed the result into the next layer of the discriminator. We compute these minibatch features separately for samples from the generator and from the training data. As before, the discriminator is still required to output a single number for each example indicating how likely it is to come from the training data: The task of the discriminator is thus effectively still to classify single examples as real data or generated data, but it is now able to use the other examples in the minibatch as \emph{side information}. Minibatch discrimination allows us to generate visually appealing samples very quickly, and in this regard it is superior to feature matching (\secc{expr}). Interestingly, however, feature matching was found to work much better if the goal is to obtain a strong classifier using the approach to semi-supervised learning described in \secc{feat_match_experiments}.

\subsection{Historical averaging}
When applying this technique, we modify each player's cost to include
a term $|| \vtheta - \frac{1}{t} \sum_{i=1}^t \vtheta[i] ||^2$,
where $\vtheta[i]$ is the value of the parameters at past time $i$.
The historical average of the parameters can be updated in an online fashion
so this learning rule scales well to long time series.
This approach is loosely inspired by the fictitious play \cite{brown1951iterative}
algorithm that can find equilibria in other kinds of games.
We found that our approach was able to find equilibria of low-dimensional,
continuous non-convex games, such as the minimax game with one player controlling 
$x$, the other player controlling $y$, and value function $(f(x) - 1) (y - 1)$,
where $f(x) = x$ for $x < 0 $ and $f(x) = x^2$ otherwise.
For these same toy games, gradient descent fails by going into extended orbits
that do not approach the equilibrium point.

\subsection{One-sided label smoothing}
Label smoothing, a technique from the 1980s
recently independently re-discovered by Szegedy et. al \cite{Szegedy-et-al-2015},
replaces the $0$ and $1$ targets for a classifier with smoothed values, like $.9$
or $.1$, and was recently shown to reduce the vulnerability of neural networks
to adversarial examples \cite{wardefarley2016}.

Replacing positive classification targets with $\alpha$ and negative targets with $\beta$, the optimal discriminator becomes
$D(\vx) = \frac{ \alpha \pdata(\vx) + \beta \pmodel(\vx) } { \pdata(\vx) + \pmodel(\vx) }$.
The presence of $\pmodel$ in the numerator is problematic because, in areas where
$\pdata$ is approximately zero and $\pmodel$ is large, erroneous samples from
$\pmodel$ have no incentive to move nearer to the data.
We therefore smooth {\em only} the positive labels to $\alpha$, leaving negative
labels set to 0.
%Note that this means the equilibrium-seeking generator cost of Eq. \ref{eq:equilibrium_seeking}
%must be updated to 
%\[
%J^{(G)}(\vtheta^{(D)}, \vtheta^{(G)}) = \KL(\frac{\alpha}{2} \Vert D(G(\vz))).
%\]

\subsection{Virtual batch normalization}
Batch normalization greatly improves
optimization of neural networks, and was shown to be highly effective for
DCGANs \cite{radford2015unsupervised}.
However, it causes the output of a neural network for an input example $\vx$ to be
highly dependent on several other inputs $\vx'$ in the same minibatch.
To avoid this problem we introduce {\em virtual batch normalization} (VBN),
in which each example $\vx$ is normalized based on the statistics collected
on a {\em reference batch} of examples that are chosen once and fixed at the
start of training, and on $\vx$ itself.
The reference batch is normalized using only its own statistics.
VBN is computationally expensive because it requires running forward propagation on two minibatches
of data, so we use it only in the generator network. %where it is most necessary, 

\section{Assessment of image quality}
\label{sec:qual}
\begin{wrapfigure}[11]{l}{0.4\textwidth}
    \fbox{
    \begin{minipage}[b][1.6cm][t]{0.41\linewidth}
	    \includegraphics[width=\textwidth]{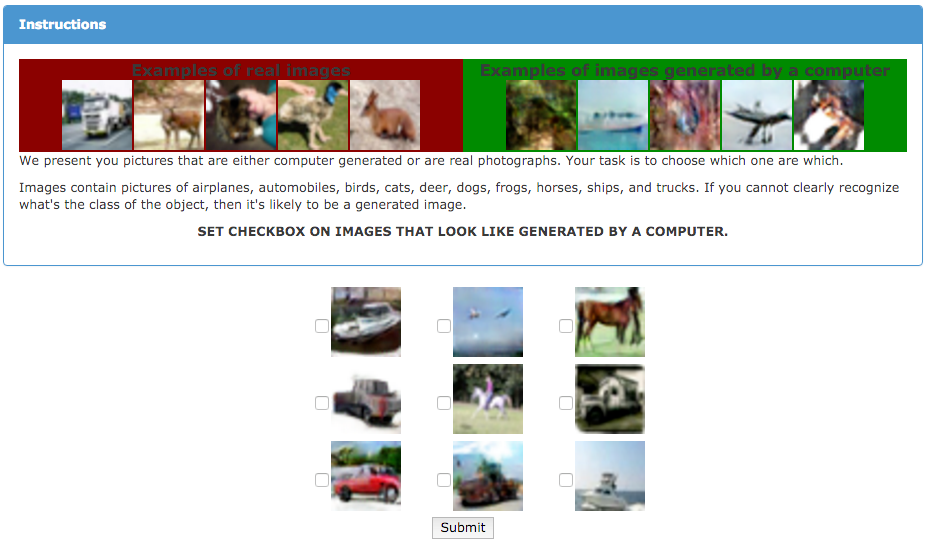}
    \end{minipage}
    }
    \hfill
    \fbox{
    \begin{minipage}[b][1.6cm][t]{0.41\linewidth}
	    \includegraphics[width=\textwidth]{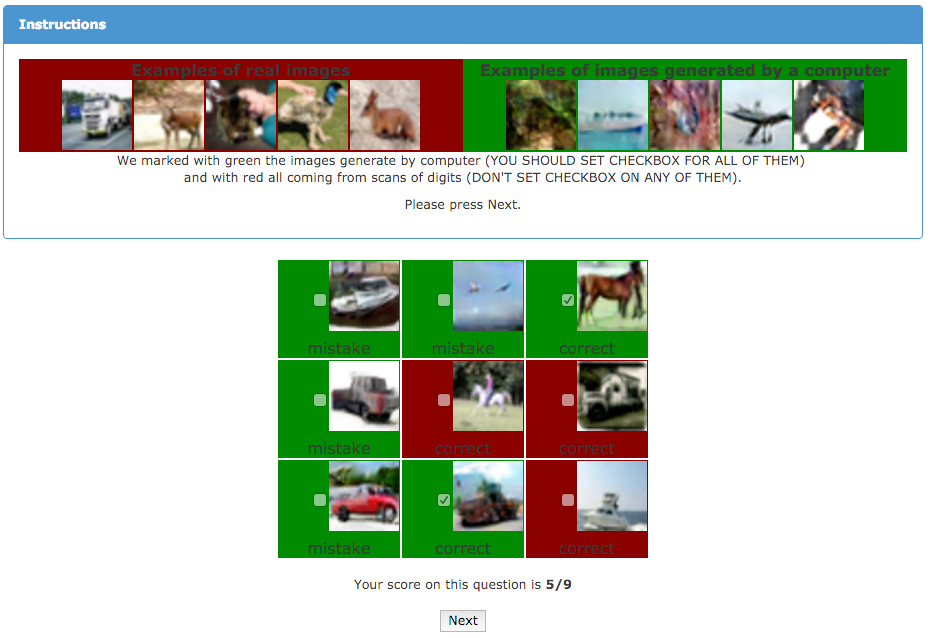}
    \end{minipage}
    }
	\caption{Web interface given to annotators. Annotators are asked to distinguish computer generated images from real ones.}
\label{fig:turing}
\end{wrapfigure}

Generative adversarial networks lack an objective function, which makes it difficult to compare performance of different models. One intuitive metric of performance can be obtained by having human annotators judge the visual quality of samples~\cite{denton2015deep}.
We automate this process using Amazon Mechanical Turk (MTurk),
using the web interface in figure~\fig{turing} (live at \url{http://infinite-chamber-35121.herokuapp.com/cifar-minibatch/}), which we use to ask annotators to distinguish between generated data and real data. The resulting quality assessments of our models are described in \secc{expr}.

A downside of using human annotators is that the metric varies depending on the setup of the task and the motivation of the annotators. We also find that results change drastically when we give annotators feedback about their mistakes: By learning from such feedback, annotators are better able to point out the flaws in generated images, giving a more pessimistic quality assessment. The left column of \fig{turing} presents a screen from the annotation process, while the right column shows how we inform annotators about their mistakes.

As an alternative to human annotators, we propose an automatic method to evaluate samples, which we find to correlate well with human evaluation: We apply the Inception model\footnote{We use the pretrained Inception model from \url{http://download.tensorflow.org/models/image/imagenet/inception-2015-12-05.tgz}. Code to compute the Inception score with this model will be made available by the time of publication.} \cite{szegedy2015rethinking} to every generated image to get the conditional label distribution $p(y | \vx)$. Images that contain meaningful objects should have a conditional label distribution $p(y | \vx)$ with low entropy. Moreover, we expect the model to generate varied images, so the marginal $\int p(y | \vx=G(z)) dz$ should have high entropy. Combining these two requirements, the metric that we propose is: $\exp(\mathbb{E}_\vx \text{KL}(p(y | \vx) || p(y)))$, where we exponentiate results so the values are easier to compare.
Our {\em Inception score} is closely related to the objective used for training generative models in CatGAN \cite{springenberg2015unsupervised}: Although we had less success using such an objective for training, we find it is a good metric for evaluation that correlates very well with human judgment. We find that it's important to evaluate the metric on a large enough number of samples (i.e. $50k$) as part of this metric measures diversity.

\section{Semi-supervised learning}
\label{sec:feat_match_experiments}
Consider a standard classifier for classifying a data point $\vx$ into one of $K$ possible classes. Such a model takes in $\vx$ as input and outputs a $K$-dimensional vector of logits $\{l_1, \dots, l_K\}$, that can be turned into class probabilities by applying the softmax: $p_\text{model}(y=j| \vx) = \frac{\exp(l_j)}{\sum_{k=1}^{K} \exp(l_k)}$.
In supervised learning, such a model is then trained by minimizing the cross-entropy between the observed labels and the model predictive distribution $p_\text{model}(y|\vx)$.

We can do semi-supervised learning with any standard classifier by simply adding
samples from the GAN generator $G$ to our data set, labeling them with a new
``generated'' class $y=K+1$, and correspondingly increasing the dimension of
our classifier output from $K$ to $K+1$.
We may then use $\pmodel(y=K+1 \mid \vx)$ to supply the probability that $\vx$
is fake, corresponding to $1-D(\vx)$ in the original GAN framework.
%If the classifier output for this class before the softmax is $l_{K+1}(x_{i})$, the estimated probability that $x_{i}$ is generated rather than real (1-$D$ in GAN) is then given by $p_\text{model}(y=K+1| x_{i}) = \frac{\exp(l_{K+1}(x_{i}))}{\sum_{k=1}^{K+1} \exp(l_k)}$, 
%just as for the other classes in the normal supervised case.
%We can now do semi-supervised learning by letting our classifier maximize the log likelihood of all observed data, maximizing $\log p_\text{model}(y_{i}=j| x_{i})$ for both labeled data and generated data. In addition,
We can now also learn from unlabeled data, as long as we know that it corresponds to one of the $K$ classes of real data by maximizing $\log p_\text{model}(y \in \{1,\ldots,K\} | \vx)$.
% = \log[1-p_\text{model}(y_{i}=K+1 | x_{i})]$ ($\log D$ in GAN).
Assuming half of our data set consists of real data and half of it is generated (this is arbitrary), our loss function for training the classifier then becomes
\begin{align*}
  L &= -\E_{\vx,y \sim p_{\text{data}}(\vx,y)}[ \log p_\text{model}(y | \vx) ] - \E_{\vx \sim G}[ \log p_\text{model}(y=K+1 | \vx) ] \\
	  &= L_{\text{supervised}} + L_{\text{unsupervised}}, \text{ where} \\
  L_{\text{supervised}} &= -\E_{\vx,y \sim p_{\text{data}}(\vx,y)}\log p_\text{model}(y | \vx, y<K+1) \\
  L_{\text{unsupervised}} &= -\{\E_{\vx \sim p_{\text{data}}(\vx)}\log[1-p_\text{model}(y=K+1| \vx)] + \E_{\vx \sim G}\log[p_\text{model}(y=K+1| \vx)]\}, \\
\end{align*}
where we have decomposed the total cross-entropy loss into our standard supervised loss function $L_{\text{supervised}}$ (the negative log probability of the label, given that the data is real) and an unsupervised loss $L_{\text{unsupervised}}$ which is in fact the standard GAN game-value as becomes evident when we substitute $D(\vx) = 1-p_\text{model}(y=K+1| \vx)$ into the expression:
 \begin{align*}
  L_{\text{unsupervised}} &= -\{ \E_{\vx \sim p_{\text{data}}(\vx)}\log D(\vx) + \E_{z \sim \text{noise}}\log(1-D(G(\vz))) \}. \\
\end{align*}
The optimal solution for minimizing both $L_{\text{supervised}}$ and $L_{\text{unsupervised}}$ is to have $\exp[l_{j}(\vx)] = c(\vx)p(y{=}j,\vx) \forall j{<}K{+}1$ and $\exp[l_{K+1}(\vx)] = c(\vx)p_{G}(\vx)$ for some undetermined scaling function $c(\vx)$. The unsupervised loss is thus consistent with the supervised loss in the sense of Sutskever et al.~\cite{sutskever2015towards}, and we can hope to better estimate this optimal solution from the data by minimizing these two loss functions jointly. In practice, $L_{\text{unsupervised}}$ will only help if it is not trivial to minimize for our classifier and we thus need to train $G$ to approximate the data distribution. One way to do this is by training $G$ to minimize the GAN game-value, using the discriminator $D$ defined by our classifier. This approach introduces an interaction between $G$ and our classifier that we do not fully understand yet, but empirically we find that optimizing $G$ using feature matching GAN works very well for semi-supervised learning, while training $G$ using GAN with minibatch discrimination does not work at all. Here we present our empirical results using this approach; developing a full theoretical understanding of the interaction between $D$ and $G$ using this approach is left for future work.

Finally, note that our classifier with $K+1$ outputs is over-parameterized: subtracting a general function $f(\vx)$ from each output logit, i.e.\ setting $l_{j}(\vx) \leftarrow l_{j}(\vx)-f(\vx) \forall j$, does not change the output of the softmax. This means we may equivalently fix $l_{K+1}(\vx)=0 \forall \vx$, in which case $L_{\text{supervised}}$ becomes the standard supervised loss function of our original classifier with K classes, and our discriminator $D$ is given by $D(\vx) = \frac{Z(\vx)}{Z(\vx) + 1}, \text{ where } Z(\vx) = \sum_{k=1}^{K} \exp[l_k(\vx)]$.

\subsection{Importance of labels for image quality}
Besides achieving state-of-the-art results in semi-supervised learning, the approach described above also has the surprising effect of improving the quality of generated images as judged by human annotators. The reason appears to be that the human visual system is strongly attuned to image statistics that can help infer what class of object an image represents, while it is presumably less sensitive to local statistics that are less important for interpretation of the image. This is supported by the high correlation we find between the quality reported by human annotators and the \emph{Inception score} we developed in \secc{qual}, which is explicitly constructed to measure the ``objectness'' of a generated image. By having the discriminator $D$ classify the object shown in the image, we bias it to develop an internal representation that puts emphasis on the same features humans emphasize. This effect can be understood as a method for transfer learning, and could potentially be applied much more broadly. We leave further exploration of this possibility for future work.

\section{Experiments}
\label{sec:expr}

We performed semi-supervised experiments on MNIST, CIFAR-10 and SVHN, and sample generation experiments on
MNIST, CIFAR-10, SVHN and ImageNet.
We provide code to reproduce the majority of our experiments.

\subsection{MNIST}
\label{sec:mnist}

\begin{wrapfigure}[17]{r}{0.5\textwidth}
	\centering
	\includegraphics[width=0.23\textwidth]{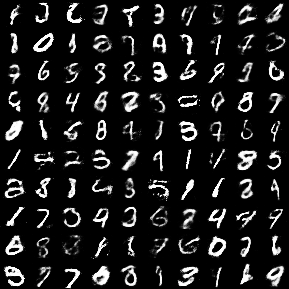}
  \hfill
	\includegraphics[width=0.23\textwidth]{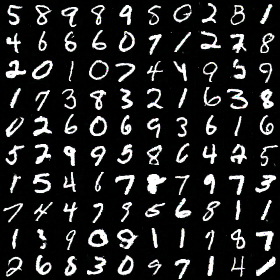}
	\caption{{\em (Left)} samples generated by model during semi-supervised training. Samples can be clearly
    distinguished from images coming from MNIST dataset. {\em (Right)} Samples generated with minibatch discrimination.
    Samples are completely indistinguishable from dataset images.}
\label{fig:mnist}
\end{wrapfigure}

The MNIST dataset contains $60,000$ labeled images of digits. We perform semi-supervised training
with a small randomly picked fraction of these, considering setups with $20$, $50$, $100$, and $200$ labeled examples.
Results are averaged over $10$ random subsets of labeled data, each chosen to have a balanced number of examples from each class. The remaining training images are provided without labels. Our networks have 5 hidden layers each.
We use weight normalization \cite{salimans2016weight} and add Gaussian noise to the output of each layer of the discriminator. \tab{mnist} summarizes our results.

Samples generated by the generator during semi-supervised learning using
feature matching (\secc{feature}) do not look visually appealing
(left \fig{mnist}). By using minibatch discrimination instead (\secc{minibatch})
we can improve their visual quality. On MTurk,
annotators were able to distinguish samples in $52.4\%$ of cases ($2000$ votes total),
where $50\%$ would be obtained by random guessing. Similarly, researchers in our
institution were not able to find any artifacts that would allow them to
distinguish samples. However, semi-supervised learning with minibatch
discrimination does not produce as good a classifier as does feature matching.

\begin{table}[ht]
  \tiny
  \centering
  \renewcommand{\arraystretch}{1.15}
  \begin{tabular}{cCCCC}
    \hline
     Model & \multicolumn{4}{c}{Number of incorrectly predicted test examples}\\
           & \multicolumn{4}{c}{for a given number of labeled samples}\\
           & 20 & 50 & 100 & 200 \\
    \hline
    \hline
     DGN~\cite{kingma2014semi} & & & $333 \pm 14$ & \\
     Virtual Adversarial~\cite{miyato2015distributional} & & & 212 & \\
     CatGAN~\cite{springenberg2015unsupervised} & & & $191 \pm 10$ & \\
		 Skip Deep Generative Model \cite{maaloe2016auxiliary} & & & $132 \pm 7$ & \\
     Ladder network~\cite{rasmus2015semi} & & & $106 \pm 37$ & \\
		 Auxiliary Deep Generative Model \cite{maaloe2016auxiliary} & & & $96 \pm 2$ & \\
    \hline
     Our model & $1677\pm452$ & $221 \pm 136$ & $93 \pm 6.5 $ & $90\pm4.2 $ \\
     Ensemble of 10 of our models & $1134\pm445$ & $142\pm96$    & $86\pm5.6$ & $81\pm4.3$\\
    \hline
  \end{tabular}
  \caption{Number of incorrectly classified test examples for the semi-supervised setting on permutation invariant MNIST. Results are averaged over $10$ seeds.}
  \label{tab:mnist}
\end{table}

\subsection{CIFAR-10}
\label{sec:cifar}

\begin{table}[h]
  \tiny
  \centering
  \renewcommand{\arraystretch}{1.15}
  \begin{tabular}{cCCCC}
    \hline
     Model & \multicolumn{4}{c}{Test error rate for}\\
           & \multicolumn{4}{c}{a given number of labeled samples}\\
           & 1000 & 2000 & 4000 & 8000 \\
    \hline
    \hline
     Ladder network~\cite{rasmus2015semi} & & & $20.40 \pm 0.47$ & \\
     CatGAN~\cite{springenberg2015unsupervised} & & & $19.58 \pm 0.46$ & \\
    \hline
     Our model& $21.83 \pm 2.01$ & $19.61 \pm 2.09$  & $18.63 \pm 2.32 $ & $17.72 \pm 1.82$\\
     Ensemble of 10 of our models & $19.22 \pm 0.54$ & $17.25 \pm 0.66$ & $15.59 \pm 0.47$ & $14.87 \pm 0.89$ \\
    \hline
  \end{tabular}
  \caption{Test error on semi-supervised CIFAR-10. Results are averaged over $10$ splits of data.}
  \label{tab:cifar}
\end{table}

CIFAR-10 is a small, well studied dataset of $32 \times 32$ natural images. We use this data set to study semi-supervised learning, as well as to examine the visual quality of samples that can be achieved. For the discriminator in our GAN we use a $9$ layer deep convolutional network with dropout and weight normalization. The generator is a 4 layer deep CNN with batch normalization. \tab{cifar} summarizes our results on the semi-supervised learning task.

%\begin{wrapfigure}[15]{l}{0.5\textwidth}
\begin{figure}
	\centering
	\includegraphics[width=0.45\textwidth]{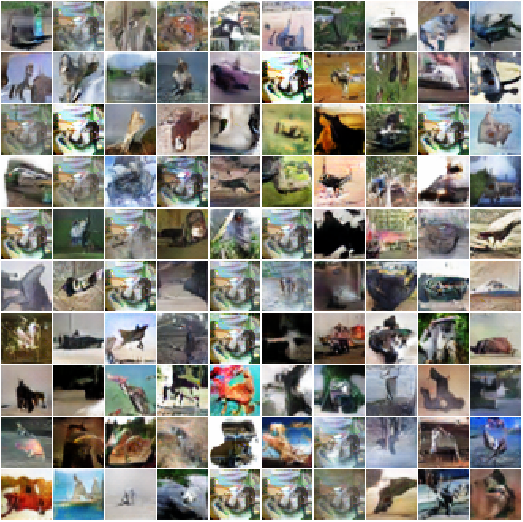}
  \hfill
	\includegraphics[width=0.45\textwidth]{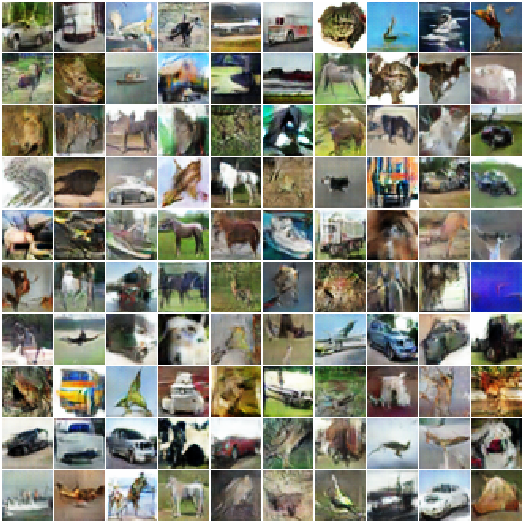}
	\caption{Samples generated during semi-supervised training on CIFAR-10 with feature matching (\secc{feature}, {\em left}) and minibatch discrimination (\secc{minibatch}, {\em right}).}
\label{fig:cifar10}
\end{figure}
%\end{wrapfigure}

When presented with $50\%$ real and $50\%$ fake data generated by our
best CIFAR-10 model, MTurk users correctly
categorized $78.7\%$ of images correctly.
However, MTurk users may not be sufficiently familiar with CIFAR-10 images
or sufficiently motivated; we ourselves were able to categorize images
with $>95\%$ accuracy.
We validated the Inception score described above by observing that MTurk
accuracy drops to $71.4\%$ when the data is filtered by using only the
top $1\%$ of samples according to the Inception score.
We performed a series of ablation experiments to demonstrate that our proposed
techniques improve the Inception score, presented in \tab{measure}.
We also present images for these ablation experiments---in our opinion, the
Inception score correlates well with our subjective judgment of image quality.
Samples from the dataset achieve the
highest value.
All
the models that even partially collapse have relatively low scores.
We caution that the Inception score should be used as a rough guide to
evaluate models that were trained via some independent criterion; directly
optimizing Inception score will lead to the generation of
adversarial examples~\cite{szegedy2013intriguing}.

\begin{table}[h]
  \tiny
  \centering
  \renewcommand{\arraystretch}{1.15}
  \begin{tabular}{p{1.5cm}CCCCCCC}
    \hline
     Samples & \includegraphics[width=\cifarwidth]{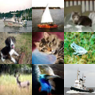} & \includegraphics[width=\cifarwidth]{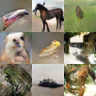} & \includegraphics[width=\cifarwidth]{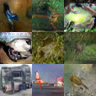}  &  \includegraphics[width=\cifarwidth]{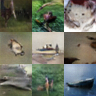} & \includegraphics[width=\cifarwidth]{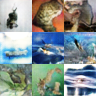} & \includegraphics[width=\cifarwidth]{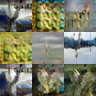} & \includegraphics[width=\cifarwidth]{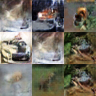}  \\
    \hline
     Model & Real data & Our methods & -VBN+BN & -L+HA & -LS & -L & -MBF \\
    \hline
     Score $\pm$ std. & $11.24 \pm .12 $ & $8.09 \pm .07$  & $7.54 \pm .07$ & $6.86 \pm .06$ & $6.83 \pm .06$ & $4.36 \pm .04$ &  $3.87 \pm .03$ \\ 
    \hline
  \end{tabular}
  \caption{Table of Inception scores for samples generated by various models for $50,000$ images.
  Score highly correlates with human judgment, and the 
  best score is achieved for natural images.
  Models that generate collapsed samples have relatively low score. This metric allows us to avoid relying on human evaluations.
  ``Our methods'' includes all the techniques described in this work, except for feature matching and historical averaging.
  The remaining experiments are ablation experiments showing that our techniques are effective.
  ``-VBN+BN'' replaces the VBN in the generator with BN, as in DCGANs. This causes a small decrease
  in sample quality on CIFAR. VBN is more important for ImageNet.
``-L+HA'' removes the labels from the training process, and adds historical averaging to compensate.
HA makes it possible to still generate some recognizable objects. Without HA, sample quality is
considerably reduced (see "-L").
``-LS" removes label smoothing and incurs a noticeable drop in performance relative to ``our methods.''
``-MBF'' removes the minibatch features and incurs a very large drop in performance, greater even than
the drop resulting from removing the labels. Adding HA cannot prevent this problem.
% NOTE:
% our methods = "calorie"
% -VBN + BN = "circle"
% -L + HA = "green"
% -LS = "citizen"
% -L = "crunch"
% -MBF = "card"
  }
  \label{tab:measure}
\end{table}

\subsection{SVHN}
\label{sec:svhn}
For the SVHN data set, we used the same architecture and experimental setup as for CIFAR-10. %Results are presented in \fig{svhn}.
%We have used the same architecture as with CIFAR-10 to also train a generative model on SVHN. \fig{svhn} shows generated images and state-of-the-art results on the semi-supervised learning task.
\begin{figure}[!htb]
    \begin{minipage}[t]{.58\textwidth}
				\vspace{2pt}
		\tiny
  \centering
  \renewcommand{\arraystretch}{1.15}
  \begin{tabular}{cCCC}
    \hline
     Model & \multicolumn{3}{c}{Percentage of incorrectly predicted test examples}\\
           & \multicolumn{3}{c}{for a given number of labeled samples}\\
           & 500 & 1000 & 2000\\
    \hline
    \hline
     DGN~\cite{kingma2014semi} & &  $36.02 \pm 0.10$ & \\
     Virtual Adversarial~\cite{miyato2015distributional} & & $24.63$ & \\
     Auxiliary Deep Generative Model \cite{maaloe2016auxiliary} & & $22.86$ & \\
     Skip Deep Generative Model \cite{maaloe2016auxiliary} & & $16.61 \pm 0.24$ & \\	 
    \hline
     Our model & $18.44 \pm 4.8$ & $8.11 \pm 1.3$ & $6.16 \pm 0.58$ \\
     Ensemble of 10 of our models & & $5.88 \pm 1.0$ & \\
    \hline
  \end{tabular}
		\end{minipage}\hspace{0.1cm}
    \begin{minipage}[t]{0.4\textwidth}
		\vspace{0pt}
		\centering
		\centerline{\includegraphics[width=0.5\textwidth]{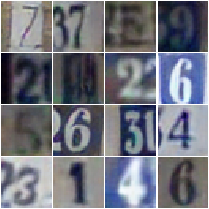}}
    \end{minipage}
		\caption{{\em (Left)} Error rate on SVHN. {\em (Right)} Samples from the generator for SVHN.}
\label{fig:svhn}
\end{figure}

\subsection{ImageNet}
\label{sec:imagenet}

\begin{figure}[htb]
	\centering
	\includegraphics[width=0.4\textwidth]{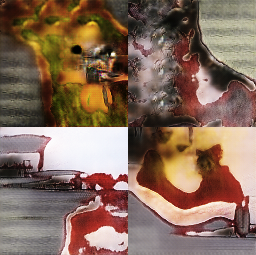}
  \hfill
	\includegraphics[width=0.4\textwidth]{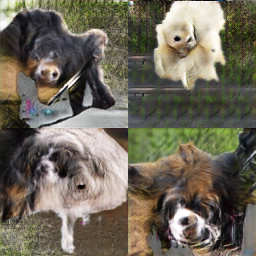}
	\caption{Samples generated from the ImageNet dataset.
  {\em (Left)} Samples generated by a DCGAN.
  {\em (Right)} Samples generated using the techniques proposed in this work.
The new techniques enable GANs to learn recognizable features of animals,
such as fur, eyes, and noses, but these features are not correctly combined
to form an animal with realistic anatomical structure.
  }
\label{fig:imagenet}
\end{figure}

We tested our techniques on a dataset of unprecedented scale: $128 \times 128$ images from the ILSVRC2012
dataset with 1,000 categories.
To our knowledge, no previous publication has applied a generative model to a dataset with both this
large of a resolution and this large a number of object classes.
The large number of object classes is particularly challenging for GANs due to their tendency to underestimate
the entropy in the distribution.
We extensively modified a publicly available implementation of DCGANs\footnote{https://github.com/carpedm20/DCGAN-tensorflow}
using TensorFlow \cite{tensorflow} to achieve high performance, using a multi-GPU implementation.
DCGANs without modification learn some basic image statistics and generate contiguous shapes
with somewhat natural color and texture but do not learn any objects.
Using the techniques described in this paper, GANs learn to generate objects
that resemble animals, but with incorrect anatomy.
Results are shown in \fig{imagenet}.

\section{Conclusion}
Generative adversarial networks are a promising class of generative models that has so far been held back by unstable training and by the lack of a proper evaluation metric. This work presents partial solutions to both of these problems. We propose several techniques to stabilize training that allow us to train models that were previously untrainable. Moreover, our proposed evaluation metric (the Inception score) gives us a basis for comparing the quality of these models. We apply our techniques to the problem of semi-supervised learning, achieving state-of-the-art results on a number of different data sets in computer vision. The contributions made in this work are of a practical nature; we hope to develop a more rigorous theoretical understanding in future work.

\small
\bibliography{bibliography}
\bibliographystyle{unsrt}

\end{document}